\documentclass{article}
\usepackage{ijcai26}

\usepackage{times}
\usepackage{soul}
\usepackage{url}
\usepackage[hidelinks]{hyperref}
\usepackage[utf8]{inputenc}
\usepackage[small]{caption}
\usepackage{amsmath}
\usepackage{amsthm}
\usepackage{amssymb}
\usepackage{booktabs}
\usepackage{multirow}
\usepackage{algorithm}
\usepackage{algorithmic}
\usepackage[switch]{lineno}
\usepackage{graphicx}

\title{Token Entropy Regularization for Multi-modal Antenna Affiliation Identification}

\author{
Dong Chen\textsuperscript{1, 2} \and
Ruoyu Li\textsuperscript{1} \and
Xinyan Zhang\textsuperscript{1} \and
Jialei Xu\textsuperscript{1} \and
Ruosen Zhao\textsuperscript{1, 3} \and
Zhikang Zhang\textsuperscript{1} \and
Lingyun Li\textsuperscript{1} \and
Zizhuang Wei\textsuperscript{1}
\affiliations{
\textsuperscript{1}Huawei \\
\textsuperscript{2}The University of Hong Kong \\
\textsuperscript{3}The Chinese University of Hong Kong, Shenzhen
}
\emails{
olichen@connect.hku.hk,
weizizhuang@huawei.com
}
}

\begin{document}

\maketitle

\begin{abstract}
     Accurate antenna affiliation identification is crucial for optimizing and maintaining communication networks. Current practice, however, relies on the cumbersome and error-prone process of manual tower inspections. We propose a novel paradigm shift that fuses video footage of base stations, antenna geometric features, and Physical Cell Identity (PCI) signals, transforming antenna affiliation identification into multi-modal classification and matching tasks. Publicly available pretrained transformers struggle with this unique task due to a lack of analogous data in the communications domain, which hampers cross-modal alignment. To address this, we introduce a dedicated training framework that aligns antenna images with corresponding PCI signals. To tackle the representation alignment challenge, we propose a novel Token Entropy Regularization module in the pretraining stage. Our experiments demonstrate that TER accelerates convergence and yields significant performance gains. Further analysis reveals that the entropy of the first token is modality-dependent. Code will be made available upon publication.
\end{abstract}

\section{Introduction}

Accurate calibration of antenna parameters is essential for network optimization but currently relies on inefficient manual tower surveys \cite{aliu2012survey,shoaib2024unveiling}. These surveys are challenging due to the large number of visually similar antennas installed at height. The Physical Cell Identity (PCI) signal offers a promising solution to this problem \cite{narmanlioglu2018mobility}. As a unique identifier in 4G and 5G networks, the PCI signals allow user devices to synchronize with a cell, thereby providing a direct digital fingerprint that links radio signals to their source antenna \cite{qi20175g,abozariba2025street}. Despite this potential, current automation efforts remain inadequate \cite{wang2015artificial}. Existing visual analysis methods fail to solve the underlying parameter-matching problem, while multi-modal AI lacks a dedicated paradigm to leverage communication signals like the PCI for this specific calibration task \cite{plested2022deep,singh2022revisiting}.

\begin{figure*}[ht]
\centering
\includegraphics[width=0.8\linewidth]{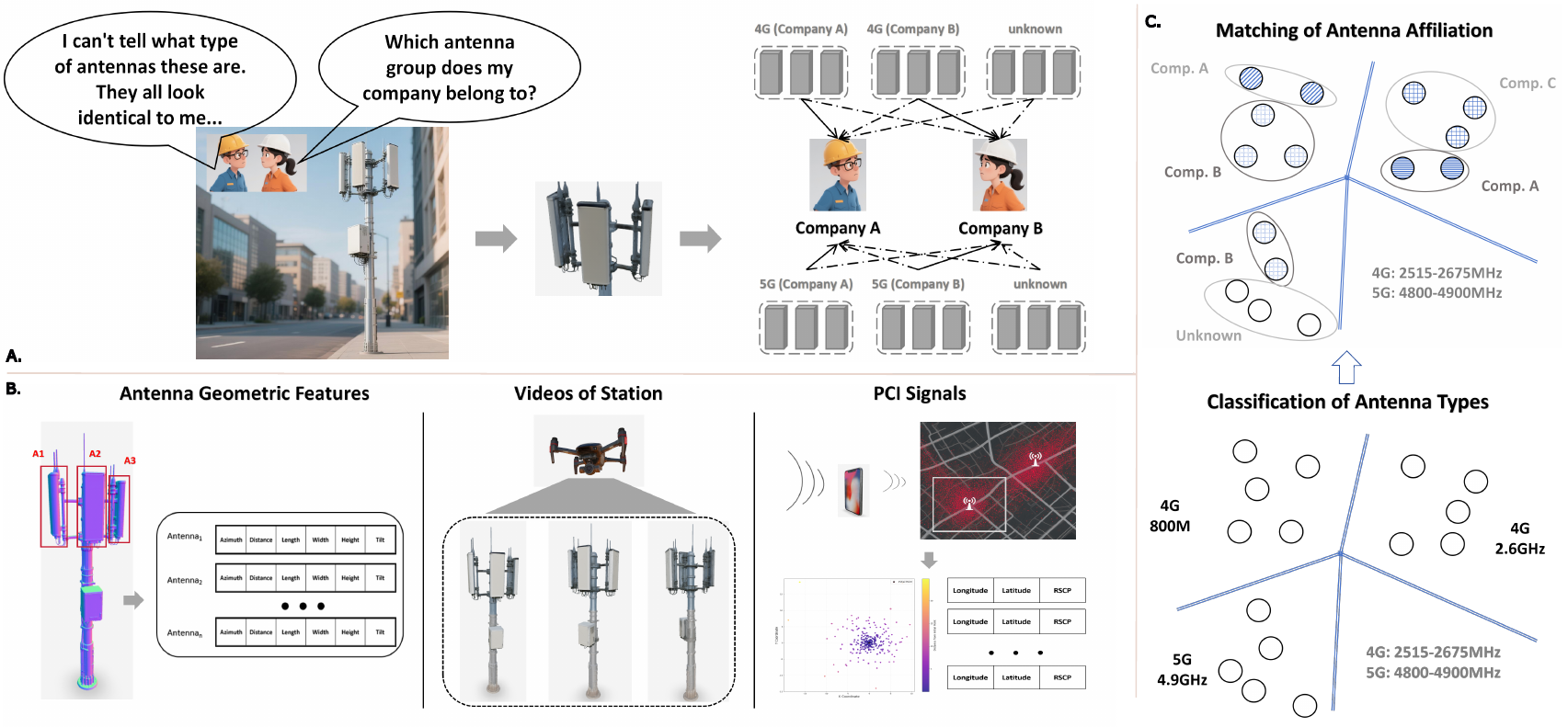}
\caption{Schematic diagram of the multi-modal antenna affiliation identification framework. (A) The practical needs of identifying antenna affiliations. (B) A multi-modal approach is proposed with base station videos, antenna geometric features, and PCI signals. (C) The task is achieved through a two-stage training process: classifying antenna types, and matching affiliations.}
\label{fig:overview}
\end{figure*}

End-to-end models often struggle to align multi-modal features, and existing pretrained large models lack specialization for communication network tasks \cite{huang2024classification,cheng2024instruction}. To address this gap, we propose a multi-modal fusion matching scheme to associate devices with their signals. Our training paradigm consists of two sequential phases. First, we employ CLIP-based contrastive learning to establish an initial alignment between visual antenna features and their corresponding PCI-level signals \cite{chen2024contrastive,liu2025selip,wang2025advancing}. Subsequently, a fine-tuning phase adapts these pretrained representations to the practical challenge of matching PCI signals to an unknown and variable set of antenna devices.

To also address representation learning with limited data, we draw inspiration from sparse rate reduction, which maximizes information gain through a coding rate function to facilitate representation learning \cite{yu2023white,hu2024depth}. This principle has been explored in the Coding Rate Reduction Transformer (CRATE), a transformer-like model that learns compressed and sparse representations \cite{yu2023white}. Building on this foundation, we introduce a Token Entropy Regularization (TER) module. TER consists of Enhanced Token Entropy (ETE) layer, which computes the entropy of token representations along the channel axis to assess encoding complexity, and a Token Entropy Loss (TEL) function, which acts as an external regularizer.

In summary, this paper has the following contributions:

(1) \textbf{A Novel Problem Formulation and Foundational Paradigm:} We are the first to formally define antenna affiliation identification as an open-world multi-modal matching task, bridging antenna classification and matching tasks. Departing from prior vision-only or generic multimodal methods, we introduce the first paradigm that integrates visual antenna data with domain-specific Physical Cell Identity (PCI) signals, establishing a new direction for data-driven network optimization.

(2)	\textbf{Introducing an Innovative Module for Feature Alignment:} To address the core challenge of aligning heterogeneous data, we propose a Token Entropy Regularization (TER) module. This framework provides a novel, theoretically-grounded method for disciplined and effective fusion of visual and signal representations, advancing multimodal learning methodology.

(3)	\textbf{A Validated, Efficient Pathway for Practical Deployment:} We demonstrate an efficient training strategy combining multimodal pretraining with targeted fine-tuning. This approach enables robust, data-efficient antenna-signal matching with substantial performance gains for 4G/5G networks, underscoring its direct viability for real-world deployment.

\begin{figure*}[t]
\centering
\includegraphics[width=0.8\linewidth]{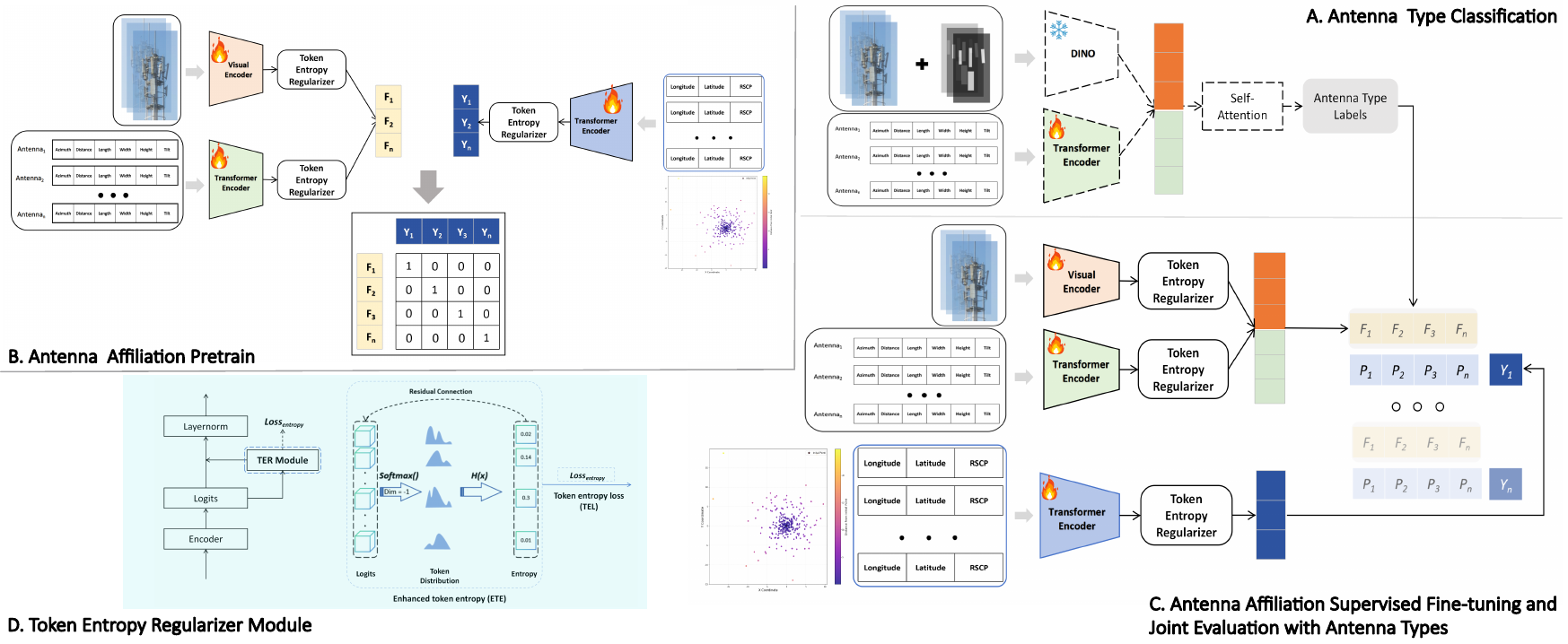}
\caption{An overview of the proposed multimodal framework for antenna affiliation identification. (A) Antenna type classification via a supervised DINOv3-Transformer. (B, C) The antenna affiliation task comprising a pretraining stage followed by supervised fine-tuning under different data-matching conditions. (D) The Token Entropy Regularization (TER) module, which encourages model convergence by regularizing the distribution of token representations.}
\label{fig:workflow}
\end{figure*}

\section{Related Work}

\subsection{Multi-modal Encoders}

The dominance of Convolutional Neural Networks (CNNs) for visual recognition was challenged by the Vision Transformer (ViT) \cite{dosovitskiy2020image}, which established the foundation by applying a pure transformer architecture directly to image patches. Subsequent work focused on improving ViT's efficiency and adaptability for vision-specific tasks. The Data-efficient Image Transformer (DeiT) \cite{touvron2021training} introduced distillation strategies to train ViTs effectively on ImageNet alone. More importantly, hierarchical architectures like the Swin Transformer \cite{liu2021swin} and the Pyramid Vision Transformer (PVT) \cite{wang2021pyramid,wang2022pvtv2} reintroduced inductive biases for multi-scale processing, enabling their use as general-purpose backbones for dense prediction. This period also saw fruitful co-evolution with modernized CNNs like ConvNeXt \cite{liu2022convnet} and alternative architectures like MLP-Mixer \cite{tolstikhin2021mlp}.

The transformer paradigm was successfully extended to video by modelling spatiotemporal relationships. Models like TimeSformer \cite{bertasius2021timesformer}, ViViT \cite{arnab2021vivit}, and Video Swin Transformer \cite{liu2022video} introduced novel attention mechanisms across frames, while hierarchical models like MViT \cite{fan2021multiscale} offered integrated multiscale video understanding.

Beyond pixels, this unified architecture proved remarkably versatile. For other modalities, such as 3D point clouds, specialized encoders like the Point Transformer \cite{zhao2021point} were developed, while the vanilla Transformer encoder \cite{vaswani2017attention} itself served as a powerful, permutation-invariant backbone for diverse structured data, from sensor signals (e.g., PCI points) to geometric features.

Two major learning paradigms on these backbones have driven progress. In self-supervised learning, masked autoencoding methods like MAE \cite{he2022masked} and knowledge distillation approaches like DINOv3 \cite{oquab2023dinov3} have learned rich, general-purpose visual features from pixels alone. In multimodal learning, contrastive models like CLIP \cite{radford2021learning} and its more efficient successor SigLIP \cite{zhai2023sigmoid} aligned visual and textual spaces, enabling zero-shot transfer. This trend towards joint representation learning has culminated in models like ImageBind \cite{girdhar2023imagebind}, which embeds multiple modalities (image, text, audio, depth, etc.) into a single, shared space, pushing the unified transformer paradigm to its logical extreme.

\subsection{Token Entropy Utilization}

Token entropy, which quantifies the information complexity or uncertainty of a token, has become a pivotal tool for analyzing and optimizing transformer-based models. It is widely studied for applications such as token selection \cite{marin2021token,wu2023tlm}, compression and pruning \cite{omri2025token,song2024less}, to improve efficiency without sacrificing performance.

Beyond efficiency, token entropy provides critical insights for model refinement. In reasoning tasks, research shows that a minority of high-entropy "forking tokens" drive most performance gains, enabling selective optimization that matches full-parameter training efficiency \cite{wang2025beyond}. For long-context modeling, frameworks like SirLLM leverage token entropy to identify and retain high-information phrases, significantly improving long-term memory in streaming dialogues without fine-tuning \cite{yao2024sirllm}.

The most advanced applications involve structured token aggregation rather than simple selection. Researches indicate that simple cluster-level token merging, which groups semantically similar tokens, often surpasses more complex selection methods that suffer from information loss or implementation overhead \cite{omri2025token,zeng2025token}. This principle aligns with findings from token merging literature, where bipartite matching approaches like ToMe demonstrate that preserving token diversity through merging is more effective than outright discarding tokens \cite{bolya2023token}.

\subsection{Sparse Rate Reduction}

Sparse rate reduction offers a principled methodology for learning sparse, discriminative, and diverse representations by maximizing the coding rate reduction \cite{yu2017compressing,hu2024depth}. Building on this perspective, white-box architectures such as CRATE  formalize the process as iterative compression where data are progressively mapped into a sparse and structured feature space \cite{yu2023white}. In parallel, recent work on token entropy regularization adopts a complementary and optimization-based view, directly shaping the output distribution by penalizing high Shannon entropy at the token level so that the model concentrates probability mass on a few confident or semantically salient tokens \cite{correia2019adaptively}.

By analogy to learned compressed sensing, this can be viewed as training the network to allocate its representational bandwidth to the most informative coordinates: token entropy regularization encourages the model to select and amplify a small subset of content-bearing tokens that are sufficient for accurate semantic reconstruction, while suppressing redundant or background tokens that contribute little to the downstream objective \cite{gao2025weight,correia2019adaptively}. The result is an adaptive, content-aware sparsity pattern in which the effective support of the representation varies with the input, much like adaptive measurement schemes in compressive imaging that focus sensing resources on salient regions, thereby improving both interpretability and efficiency without sacrificing predictive performance \cite{zhu2015adaptive}.

\section{Methodology}

Following the task formulation in section 3.1, section 3.2 details the multi-modal data acquisition process. Subsequently, Section 3.3 introduces the proposed Token Entropy Regularizer, whose role in the framework in Figure~\ref{fig:workflow}d. Finally, the overall pipeline for the antenna affiliation task is presented, which comprises antenna type classification (Figure~\ref{fig:workflow}a) and is trained using a two-stage strategy (Figure~\ref{fig:workflow}b, c). A comparison with end-to-end training is also provided, with detailed implementations available in the supplementary materials.

\subsection{Task Formulation}
The core problem addressed in this work is establishing reliable correspondence between physical antenna devices observed in the wild and their unique communication signatures, particularly the Physical Cell Identity (PCI). Traditional approaches rely on manual tower-climbing surveys, which are labour-intensive, error-prone, and impractical for large-scale networks. We redefine this practical engineering challenge as an open-world multi-modal matching problem in computer vision and signal processing.

Formally, our antenna affiliation identification task can be formulated as follows: given a set of visual observations \( \mathcal{V} = \{v_1, v_2, ..., v_M\} \) containing \( M \) antenna devices from station videos or images, and a set of PCI signals \( \mathcal{S} = \{s_1, s_2, ..., s_N\} \) detected from the same station, the objective is to learn a matching function \( f: \mathcal{V} \times \mathcal{S} \rightarrow [0,1] \) that determines whether antenna device \( v_i \) corresponds to PCI signal \( s_j \). This formulation transforms the practical antenna affiliation identification problem into a multi-modal classification and matching paradigm.

The inputs to our system comprise three complementary modalities:
\begin{itemize}
    \item \textbf{Visual data} \( \mathcal{V} \): Raw video footage or images capturing antenna installations at base station sites, from which individual antenna devices are extracted.
    \item \textbf{Geometric features} \( \mathcal{G} \): Structured representations of antenna physical attributes, including dimensions, shape characteristics, and installation parameters.
    \item \textbf{PCI signals} \( \mathcal{S} \): Digital identifiers by cellular antennas that serve as unique fingerprints for signal source attribution.
\end{itemize}

The output space constitutes an open-set matching assignment where the number of antennas \( M \) and detected PCIs \( N \) may not be equal (\( M \neq N \) generally), and the correspondence between visual devices and communication signals must be established without prior knowledge of the exact matching relationship. This open-world condition reflects real-world deployment scenarios.

The key challenges in this formulation include:
\begin{enumerate}
    \item \textbf{Modality Gap}: The significant distributional differences between visual features and signal features create alignment difficulties.
    \item \textbf{Data Scarcity}: Limited annotated multi-modal pairs for antenna-PCI correspondence, requiring data-efficient learning approaches.
    \item \textbf{Open-World Complexity}: The variable number of antennas and signals per station, with potential missing detections or false positives in both modalities.
    \item \textbf{Domain Specificity}: The absence of similar tasks in existing literature means pretrained models lack relevant inductive biases for this communication network application.
\end{enumerate}

This task formulation distinguishes itself from conventional multi-modal learning problems by addressing the unique constraints of communication infrastructure, where visual and signal modalities must be correlated under practical deployment constraints to solve a concrete industrial problem.

\subsection{Multi-modal Data Acquisition}
\label{sec:data_acquisition}

As illustrated in Figure~\ref{fig:overview}, our multi-modal dataset is constructed from three complementary sources, each providing a unique perspective on the antenna deployment and its radio characteristics.

\textbf{Visual Data.} High-resolution RGB image sequences of communication towers are captured by a drone performing a controlled orbit around each site. This procedure yields approximately 30 images per station, comprehensively documenting pole-mounted and rooftop antennas against diverse backgrounds. From these images, we extract individual antenna instances and estimate their key geometric features---azimuth ($\theta$), length ($L$), width ($W$), height ($H$), and tilt ($\alpha$)---using 3D reconstruction techniques (e.g., Structure-from-Motion). The visual modality is therefore represented as $\mathcal{V} = \{v_1, v_2, ..., v_M\}$, where $v_i$ is the feature vector encapsulating the $i$-th antenna's appearance and geometry.

\textbf{Signal Data.} Concurrently, Physical Cell Identity (PCI) signal data is collected by performing radio frequency scans on predefined 4G and 5G network frequencies. This dataset provides a set $\mathcal{S} = \{s_1, s_2, ..., s_N\}$, where each signal $s_j$ is characterized by its geographic coordinates (latitude $\phi_j$, longitude $\lambda_j$) and signal intensity metrics, such as Reference Signal Received Power (RSRP).

\textbf{Ground Truth Association.} The core challenge of cross-modal matching requires a precise, verified link between visual antennas and their emitted signals. The ground truth affiliation for antenna $v_i$ is established by associating its unique visual identifier (ID) with a corresponding PCI-level signal $s_j$, using standardized network reference data. This yields a set of verified matching pairs $\mathcal{P} = \{(v_a, s_b) \,|\, \text{antenna } v_a \text{ is affiliated with signal } s_b\}$, which serves as the supervision for our learning task.

\subsection{Token Entropy Regularizer}
\label{subsec:token_entropy}

A fundamental challenge in training multi-modal networks lies in aligning heterogeneous feature distributions. We hypothesize that dynamically measuring and regularizing the information encoding density of learned token representations can facilitate this alignment. To this end, we propose the Token Entropy Regularization (TER) module, a sparsity-inducing mechanism that encourages the model to form compact, high-fidelity representations by penalizing tokens with high entropy, those that are uniformly distributed and thus less informative.




\textbf{Enhanced Token Entropy (ETE):} Given the feature representation $\mathbf{F}_m \in \mathbb{R}^{T \times D}$ generated by the modality encoder $E_m$, where $T$ denotes the number of tokens and $D$ the feature dimension, we compute a probability distribution over the feature channels for each token. Specifically, for the $t$-th token $\mathbf{f}_{m,t} \in \mathbb{R}^D$, we apply the SoftMax function along the channel dimension to obtain a discrete probability distribution:

\begin{align}
    \mathbf{p}_{m,t} &= \text{SoftMax}(\mathbf{f}_{m,t}) = \left[\frac{\exp(f_{m,t,i})}{\sum_{j=1}^{D} \exp(f_{m,t,j})}\right]_{i=1}^{D}
\end{align}

The token entropy $H_{m,t} \equiv H(\mathbf{f}_{m,t})$ is then calculated as the Shannon entropy of this distribution, yielding a scalar measure of the token's feature complexity:

\begin{align}
    H_{m,t} &= -\sum_{i=1}^{D} p_{m,t,i} \cdot \log(p_{m,t,i})
\end{align}

A high entropy value indicates a token whose activations are evenly spread across feature channels, suggesting a less discriminative or "noisy" representation. Conversely, a low entropy value indicates a "sparse" token with activations concentrated on a few channels, typically associated with more specific, informative features.

To inject this self-awareness into the learning process, the scalar entropy $H_{m,t}$ is processed by a lightweight projection head (e.g., a linear layer followed by LayerNorm) and added back to the original token feature via a gated residual connection:

\begin{align}
    \tilde{\mathbf{f}}_{m,t} = \mathbf{f}_{m,t} + \gamma \cdot \mathrm{LayerNorm}\left( \mathbf{W} H_{m,t} + \mathbf{b} \right),
\end{align}

where $\mathbf{W} \in \mathbb{R}^{D \times 1}$ and $\mathbf{b} \in \mathbb{R}^{D}$ are learnable parameters, and $\gamma$ is a learnable scalar gate. The output of the ETE layer is the entropy-augmented feature matrix $\tilde{\mathbf{F}}_m = [\tilde{\mathbf{f}}_{m,1}, \tilde{\mathbf{f}}_{m,2}, \ldots, \tilde{\mathbf{f}}_{m,T}]^\top$.

\textbf{Token Entropy Loss (TEL).} Beyond the architectural modification, we explicitly regularize token sparsity via an auxiliary loss. The TEL encourages the model to produce low-entropy, high-confidence tokens by minimizing the mean token entropy across the batch and modalities:

\begin{align}
    \mathcal{L}_{TEL} = \frac{1}{|\mathcal{B}|} \frac{1}{T} \sum_{m \in \{\mathcal{V},\mathcal{S},\mathcal{G}\}} \sum_{t=1}^{T} H_{m,t},
\end{align}

where $\mathcal{B}$ denotes the current training batch and $|\mathcal{B}|$ is its size. The total training objective becomes a weighted sum of the primary task loss (e.g., contrastive or matching loss) and the regularization term:

\begin{align}
    \mathcal{L}_{\text{total}} = \mathcal{L}_{\text{task}} + \lambda \, \mathcal{L}_{TEL},
\end{align}

where $\lambda$ is a hyperparameter controlling the strength of the entropy regularization. This formulation enables a principled trade-off between task performance and the compactness of the learned multi-modal representations.

\section{Experiments}
\subsection{Implementation Detials}

\textbf{Hyperparameter:} All experiments were conducted on a single NVIDIA A800 GPU. We used the Adam optimizer for pretraining and AdamW for fine-tuning. The learning rate followed a cosine decay schedule, preceded by a linear warm-up phase spanning 10 epochs. Each training task ran for 200 epochs.

\textbf{Fixed Encoders:} The encoders for antenna geometric features and PCI signals are based on a modified Transformer architecture, with 4 layers, 8 attention heads, and a feature dimension of 1024.

\subsection{Antenna Type Classification}

\textbf{Vision feature extraction:} We evaluated multiple components of our multi-modal framework for antenna type classifications. For visual feature extraction, we compared DINOv3 (ViT-L/16 and ViT-H+/16 variants) with PVTv2, testing three token generation strategies: [CLS] token, top layer features (TLF) with global average pooling, and their fusion. Geometric features were encoded using relative position encoding and normalized, comparing 3D point clouds against abstract semantic representations. For multi-modal fusion, we evaluated cross-attention, self-attention, and MLP-based approaches.

As shown in Table~\ref{tab:comprehensive_results}, DINOv3 significantly outperforms PVTv2, with ViT-L/16 achieving the best results (86.4\% accuracy using TLF). Semantic geometric features (86.4\%) proved more effective than raw 3D points (84.6\%). Among fusion methods, self-attention achieved the highest accuracy (86.4\%), though differences between fusion strategies were minimal. These results demonstrate that DINOv3 with semantic geometric features and self-attention fusion provides the optimal configuration for our task.

\begin{table}[t]
  \centering
  \footnotesize
  \begin{tabular}{@{}llc@{}}
    \toprule
    \textbf{Vision Encoder} & \textbf{Variant} & \textbf{Acc. (\%)} \\
    \midrule
    PVTv2-b3 & LAYER-3 & 67.5 \\
    PVTv2-b3 & LAYER-ALL & 68.8 \\
    ViT-H+/16 & [CLS] & 75.5 \\
    ViT-H+/16 & [CLS]+TLF & 74.9 \\
    ViT-H+/16 & TLF & 76.6 \\
    ViT-L/16 & [CLS] & 85.9 \\
    ViT-L/16 & [CLS]+TLF & 85.5 \\
    ViT-L/16 & \ TLF & \underline{86.4} \\
    \midrule
    \textbf{Geometric Features} & & \\
    \cmidrule{1-2}
    3D Points & - & 84.6 \\
    Semantic Geometric & - & \underline{86.4} \\
    \midrule
    \textbf{Fusion Method} & & \\
    \cmidrule{1-2}
    Cross-Attention & - & 85.9 \\
    MLP & - & 86.1 \\
    Self-Attention & - & \underline{86.4} \\
    \bottomrule
  \end{tabular}
  \caption{Comprehensive evaluation of vision encoders, geometric features, and fusion methods for antenna type classification. Best: DINOv3 (ViT-L/16) with TLF, semantic geometric encoding, and self-attention fusion.}
  \label{tab:comprehensive_results}
\end{table}

\subsection{Antenna Affiliation Pretraining }

We evaluate the contribution of the proposed Token Entropy Regularization (TER) module by conducting comparative experiments with three modified visual encoders: TimeSformer, Video Swin Transformer, and ViViT. As shown in Table~\ref{tab:detailed_pretrain}, the TER module consistently improved the accuracy of all three architectures, demonstrating that its mechanism for explicitly modeling token complexity effectively enhances feature representations. The Video Swin Transformer achieved the highest performance, with a top-1 accuracy of 65.12\% when full TER regularization was applied.

\begin{table}[H]
\centering
\small
\begin{tabular}{lcccccc}
\toprule
\textbf{Vision Encoder} & \textbf{Config} & \textbf{Acc@1} & \textbf{Acc@3} \\
\midrule
\multirow{2}{*}{TimeSformer}
& Baseline & 53.49 & 88.37 \\
& + TER & 62.79 & \underline{95.35} \\
\midrule
\multirow{2}{*}{ViViT}
& Baseline & 51.16 & 88.37 \\
& + TER & 53.49 & 90.70 \\
\midrule
\multirow{2}{*}{Video Swin}
& Baseline & 55.81 & 86.05 \\
& + TER & \underline{65.12} & 86.05 \\
\bottomrule
\end{tabular}
\caption{Comprehensive evaluation of token entropy regularizer module across different visual encoders in pretraining.}
\label{tab:detailed_pretrain}
\end{table}

\subsection{Integrated Antenna Affiliation Evaluation}

We evaluated the practical efficacy of our method by comparing two training paradigms: End-to-End (End2End) learning and supervised fine-tuning on the pretrained model (Pretrain + SFT). Results on real-world 4G/5G antenna-PCI matching data (Table~\ref{tab:sft_results}) show that the Pretrain + SFT paradigm substantially and consistently outperforms the End2End approach across all three video encoder architectures. This universal improvement, with top-1 accuracy gains ranging from approximately +26\% (Video Swin) to +38\%(TimeSformer), validates the critical advantage of the proposed pretraining stage. Under this superior paradigm, the Video Swin encoder achieves the best performance, with an overall top-1 accuracy of 87.91\% and a top-3 accuracy of 91.94\%.

\begin{table*}[t]
\centering
\small 
\begin{tabular}{l l c c c c c c}
\toprule
 & & \multicolumn{2}{c}{\textbf{4G Accuracy (\%)}} & \multicolumn{2}{c}{\textbf{5G Accuracy (\%)}} & \multicolumn{2}{c}{\textbf{Overall}} \\
\cmidrule(lr){3-4} \cmidrule(lr){5-6} \cmidrule(lr){7-8}
\textbf{Video Encoder} & \textbf{Training Paradigm} & \textbf{@1} & \textbf{@3} & \textbf{@1} & \textbf{@3} & \textbf{@1} & \textbf{@3} \\
\midrule
TimeSformer & End2End & 30.43 & 65.22 & 30.30 & 63.64 & 30.36 & 64.29 \\
ViViT & End2End & \underline{52.17} & \underline{86.96} & 60.61 & 84.85 & 56.39 & 85.91 \\
Video Swin & End2End & 51.52 & 86.96 & \underline{72.73} & \underline{90.91} & \underline{62.12} & \underline{89.93} \\
\midrule
TimeSformer & Pretrain + SFT & 69.56 & 86.95 & 66.66 & 74.07 & 68.11 & 80.51 \\
ViViT & Pretrain + SFT & 78.26 & 91.30 & 81.48 & 85.18 & 79.87 & 88.24 \\
Video Swin & Pretrain + SFT & \underline{86.95} & \underline{91.30} & \underline{88.88} & \underline{92.59} & \underline{87.91} & \underline{91.94} \\
\bottomrule
\end{tabular}
\caption{Comprehensive evaluation of antenna-PCI matching performance across video encoders and training paradigms, showing both @1 and @3 accuracy metrics. TER framework consistently improves performance across all metrics, with ViViT achieving optimal results.}
\label{tab:sft_results}
\end{table*}

\subsection{Visualization}

The following visualizations provide qualitative insights into model performance for the antenna type classification and matching tasks.

\begin{figure}[ht]
\centering
\includegraphics[width=0.8\linewidth]{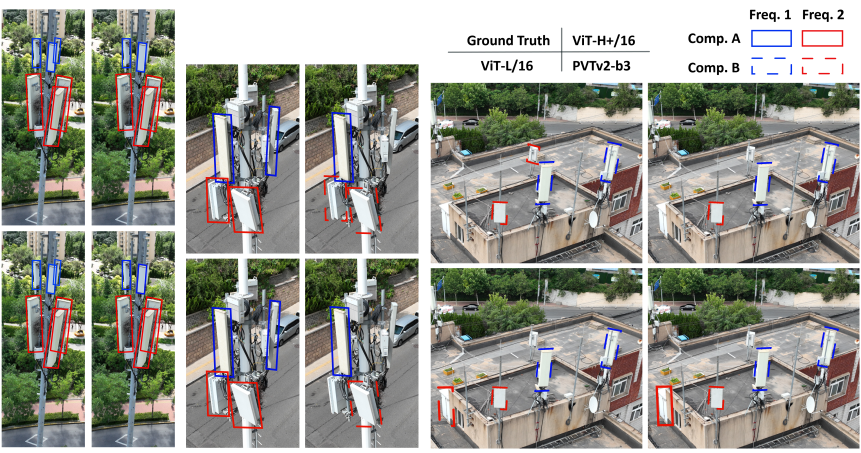}
\caption{Visual comparison of antenna type classification. Rows show different antenna instances; columns compare ground truth (left) with predictions from three distinct models. Line style indicates manufacturer (solid: Company A, dashed: Company B); color indicates frequency band (blue: low, red: high).}
\label{fig:sup_1}
\end{figure}

Figure~\ref{fig:sup_1} compares the antenna type classification results of three different models against ground truth annotations. The visualization uses a consistent encoding scheme: solid and dashed lines represent antennas from Company A and Company B, respectively, while blue and red colors denote low-frequency and high-frequency variants. This structured comparison reveals systematic differences in classification accuracy across models.

\begin{figure}[ht]
\centering
\includegraphics[width=0.8\linewidth]{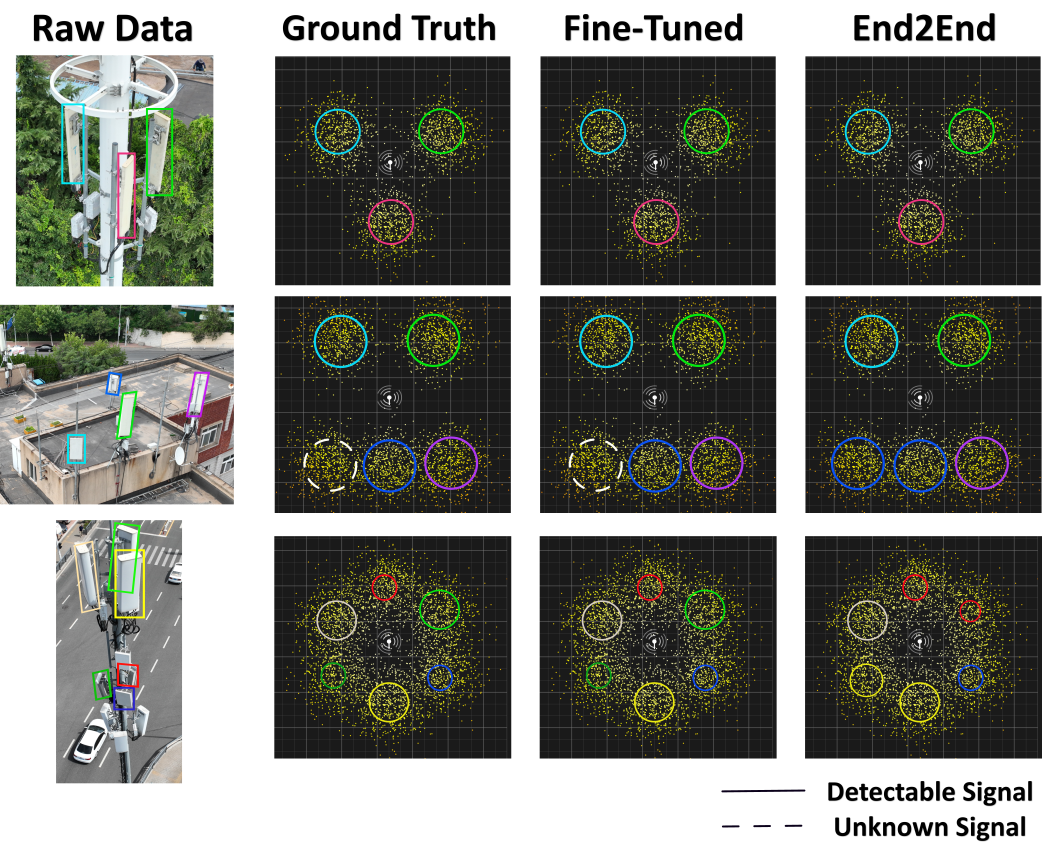}
\caption{Antenna affiliation matching results. Each row presents a distinct urban scenario with communication infrastructure. Yellow points mark detected PCI signals, with solid circles indicating verified matches and dashed circles indicating unclassified signals. Colors represent different signal classes.}
\label{fig:sup_2}
\end{figure}

Figure~\ref{fig:sup_2} visualizes antenna affiliation matching performance in complex urban environments. Detected PCI signals (yellow points) are overlaid on scenario backgrounds, with solid circles representing correctly matched signals and dashed circles indicating unclassified or ambiguous detections. Circle colors correspond to specific signal classes. The visualization demonstrates the model's capability to maintain accurate signal-source associations across challenging real-world conditions.

\subsection{Ablation Study}

\begin{figure}[H]
\centering
\includegraphics[width=0.8\linewidth]{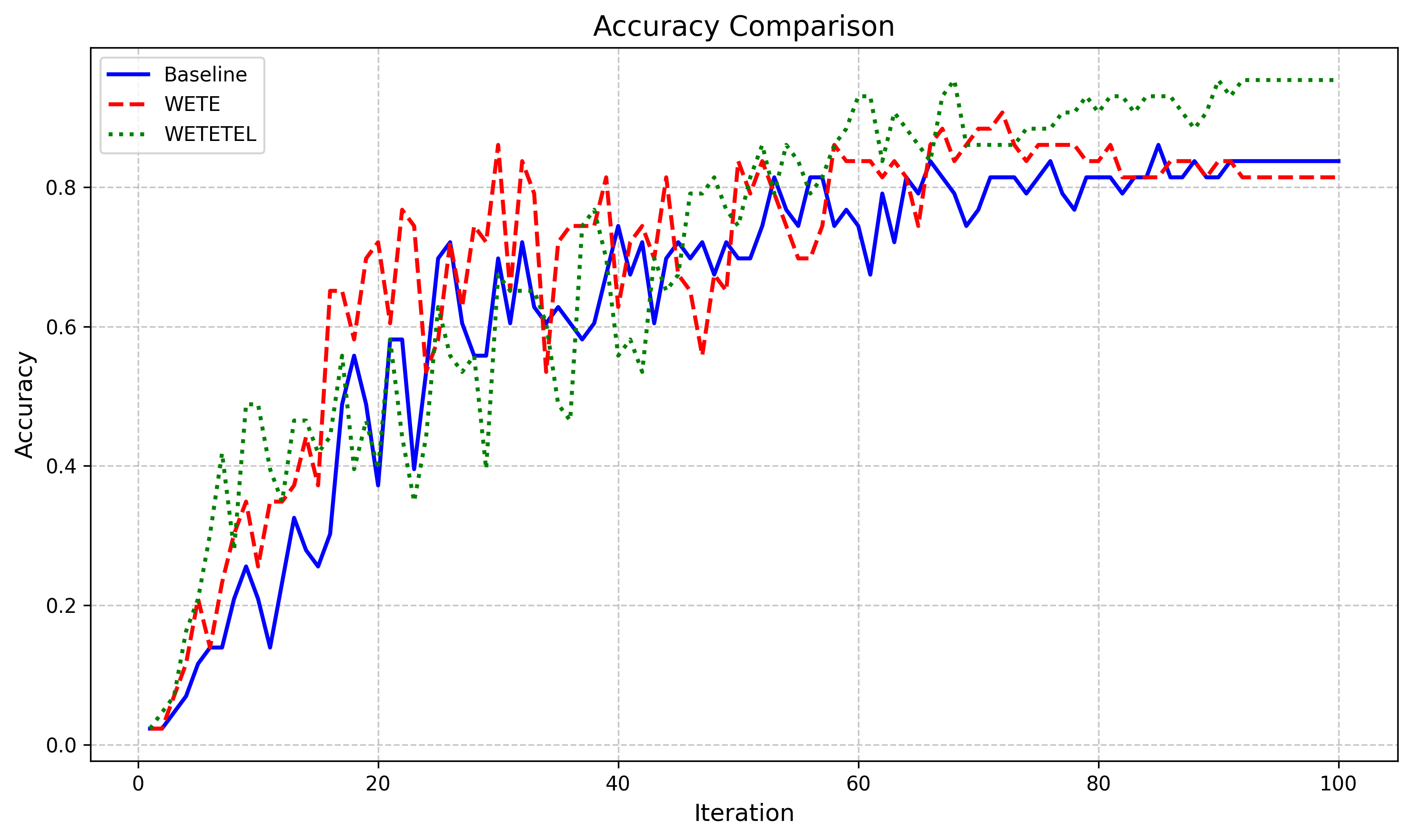}
\caption{Accuracy during pretraining. Our proposed ETE module converges faster and to a higher accuracy than the baseline. Performance is further improved by the full TER method (ETE + TEL).}
\label{fig:pretrain_curves}
\end{figure}

\textbf{Pretraining stage:}  As shown in Figure~\ref{fig:pretrain_curves}, the ETE module (red) accelerates convergence and achieves higher final accuracy compared to the baseline (blue). The complete TER method, which incorporates the TEL constraint (green), delivers the best performance. This indicates that token-level complexity perception enables the model to learn more effective and adaptive feature representations.

\begin{table*}[t!]
\centering
\small
\setlength{\tabcolsep}{5pt} 
\begin{tabular}{lcccccc}
\hline
\textbf{Method} & \textbf{i2t\_R@1} & \textbf{i2t\_R@5} & \textbf{i2t\_R@10} & \textbf{t2i\_R@1} & \textbf{t2i\_R@5} & \textbf{t2i\_R@10} \\
\hline
\addlinespace
\multicolumn{7}{l}{\textbf{CLS Setting}} \\
Baseline & 0.5040 & 0.7986 & 0.8705 & 0.4797 & 0.7991 & 0.8700 \\
+ TER & 0.5027 ($-$0.25\%) & 0.8023 ($+$0.46\%) & 0.8737 ($+$0.37\%) & 0.4790 ($-$0.15\%) & 0.7958 ($-$0.40\%) & 0.8697 ($-$0.03\%) \\
\hline
\addlinespace
\multicolumn{7}{l}{\textbf{Non-CLS Setting}} \\
Baseline & 0.5077 & 0.7949 & 0.8737 & 0.4869 & 0.8013 & 0.8730 \\
+ TER & 0.5121 (+0.88\%) & 0.8114 (+2.08\%) & 0.8791 (+0.62\%) & 0.4998 (+2.64\%) & 0.8080 (+0.83\%) & 0.8772 (+0.48\%) \\
\hline
\end{tabular}
\caption{Performance comparison on Flickr8K dataset with and without TER regularization. Values represent recall metrics with improvement percentages shown in parentheses.}
\label{tab:flickr8k_results}
\end{table*}

\begin{figure}[t]
\centering
\includegraphics[width=0.8\linewidth]{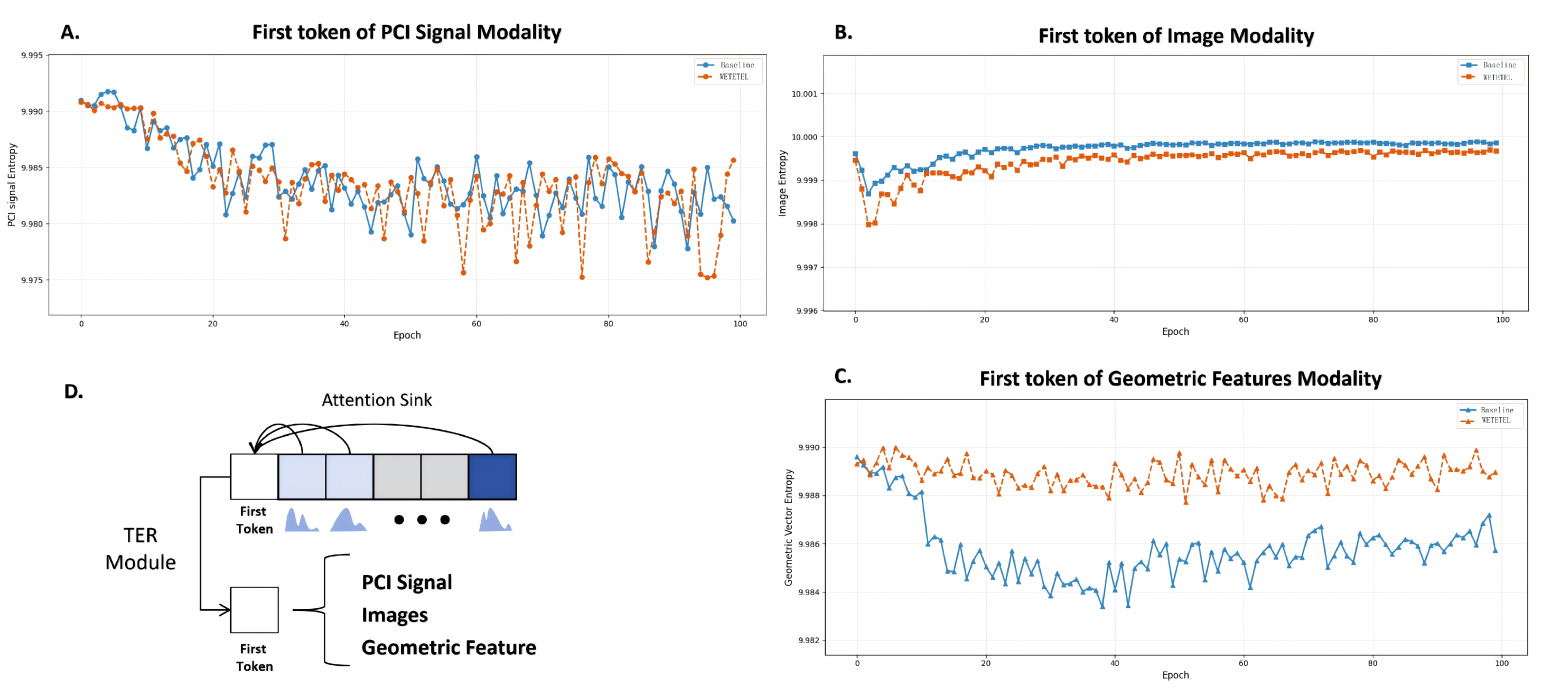}
\caption{Dynamics of first-token entropy under regularization settings. (A-C) Entropy trajectories for the PCI signal, image, and geometric feature modalities, illustrating the adaptive sparsification effect of the TER module during pretraining. (D) Schematic representation of the first token in the token entropy regularization framework.}
\label{fig:token_entropy}
\end{figure}

\textbf{Analysis of First Token Entropy in Pretraining:} To investigate how the proposed TER module contributes to the adaptive sparse rate encoding, we analyzed the evolution of first token entropy for each modality during pretraining. The first token is selected due to the attention sink property of current attention mechanisms, as shown in Figure~\ref{fig:token_entropy} (D), making it particularly informative for measuring representation complexity \cite{xiao2023streamingllm}. As shown in Figure~\ref{fig:token_entropy} (A-C), we tracked first token entropy for each modality (PCI signal, image, geometric vector) over 100 epochs under different regularization conditions.

The analysis yields two principal findings that align with the goal of learning sparse, informative representations. First, the significant disparity in baseline entropy across modalities underscores their inherent differences in representation complexity, highlighting the need for an adaptive sparsification strategy. Second, the figure demonstrates a clear, modality-dependent effect on entropy modulation, effectively achieving the "adaptive" sparsity desired by compressing complex representations while preserving essential information.

\subsection{Token Entropy Regularizer Generalization}

To evaluate the generalization capability of the proposed TER method, we conducted 100-epoch training experiments on the Flickr8K dataset under two distinct settings: CLS token-based representation and non-CLS token representation. The results demonstrate the adaptive sparsification effect of our approach across different architectural configurations. Quantitative results are summarized in Table~\ref{tab:flickr8k_results}.

\textbf{CLS Token Usage:} Under the CLS token setting, the TER-regularized model exhibits mixed performance compared to the baseline. While image-to-text recall at 5 (i2t\_R@5) improves by +0.46\% and i2t\_R@10 increases by +0.37\%, other metrics show minor declines, with the largest drop observed in t2i\_R@5(-0.40\%). This indicates that CLS-based representations, which aggregate global information into a single token, may not fully leverage the token-wise sparsification induced by TER.

\textbf{Non-CLS Token Usage:} In contrast, the non-CLS setting demonstrates consistent improvements across all evaluation metrics. The TER-regularized model achieves significant gains in text-to-image recall at 1 (t2i\_R@1) with an improvement of +2.64\% and image-to-text recall at 5 (i2t\_R@5) with +2.08\% improvement. This performance enhancement aligns with the core principle of our TER approach, which promotes sparse, informative representations at the token level rather than relying on aggregated CLS embeddings. The results indicate that by preserving fine-grained token-level information while applying entropy constraints, the model learns more discriminative representations for cross-modal retrieval tasks.

\section{Conclusion}

This paper introduced a novel multimodal framework for automated antenna affiliation identification, enhanced by a token-level regularizer. By synergistically fusing visual data from drone-captured imagery with 4G/5G network signals, the proposed system overcomes critical limitations of conventional manual methods. Experimental results validate the framework's efficacy, demonstrating substantial improvements from the proposed Token Entropy Regularization. These advancements provide significant practical benefits for telecommunication infrastructure management, enabling more efficient network planning, operational optimization, and asset management while mitigating the safety risks inherent in manual field verification.

\newpage
\bibliographystyle{named}
\bibliography{ijcai26}

\end{document}